\documentclass{article}

\usepackage{arxiv}

\usepackage[utf8]{inputenc} 
\usepackage[T1]{fontenc}    
\usepackage{hyperref}       
\usepackage{url}            
\usepackage{booktabs}       
\usepackage{amsfonts}       
\usepackage{nicefrac}       
\usepackage{microtype}      
\usepackage{lipsum}		
\usepackage{graphicx}
\usepackage{natbib}
\usepackage{doi}
\usepackage{amssymb,amsmath,amsthm}
\usepackage{color}
\usepackage[toc,page]{appendix}
\usepackage[ruled,vlined,linesnumbered,noresetcount]{algorithm2e}

\theoremstyle{definition}

\SetKwInput{KwInput}{Input}
\SetKwInput{KwOutput}{Output}

\title{Supervised Class-pairwise NMF for Data Representation and Classification}

\author{ \href{https://orcid.org/0000-0002-9065-9591}{Rachid Hedjam}\\
	Department of Computer Science\\
	Sultan Qaboos University\\
	Muscat, Oman \\
	\texttt{rachid.hedjam@squ.edu.om} \\
	\And
	{Abdelhamid Abdesselam} \\
Department of Computer Science\\
Sultan Qaboos University\\
Muscat, Oman \\
	\texttt{ahamid@squ.edu.om} \\
	 \AND
	  Seyed Mohammad Jafar Jalali \\
	 Institute for Intelligent Systems Research and Innovation (IISRI) \\
  Deakin University\\
  Australia\\
	 \texttt{sjalali@deakin.edu.au} 
 \AND
	 Imran Khan \\
	 Department of Computer Science\\
Sultan Qaboos University\\
Muscat, Oman \\
	\texttt{i.khan@squ.edu.om} 
 \AND
	  Samir Brahim Belhaouari \\
	Division of Information and Computing Technology\\
College of Science and Engineering\\
Hamad Bin Khalifa University\\
Doha, Qatar\\
	\texttt{sbelhaouari@hbku.edu.qa} \\
}

\renewcommand{\shorttitle}{\textit{arXiv} Template}

\begin{document}
\maketitle

\begin{abstract}
Various Non-negative Matrix factorization (NMF) based methods add new terms to the cost function to adapt the model to specific tasks, such as clustering, or to preserve some structural properties in the reduced space (e.g., local invariance). The added term is mainly weighted by a hyper-parameter to control the balance of the overall formula to guide the optimization process towards the objective. The result is a \textit{parameterized} NMF method. However, NMF method adopts unsupervised approaches to estimate the factorizing matrices. Thus, the ability to perform prediction (e.g. classification) using the new obtained features is not guaranteed. The objective of this work is to design an evolutionary framework to learn the hyper-parameter of the parameterized NMF and estimate the factorizing matrices in a supervised way to be more suitable for classification problems. Moreover, we claim that applying NMF-based algorithms separately to different class-pairs instead of applying it once to the whole dataset improves the effectiveness of the matrix factorization process.  This results in training multiple parameterized NMF algorithms with different balancing parameter values. A cross-validation combination learning framework is adopted and a Genetic Algorithm is used to identify the optimal set of hyper-parameter values. The experiments we conducted on both real and synthetic datasets demonstrated the effectiveness of the proposed approach.
\end{abstract}


\section{Introduction}
\label{sec:intro}

The NMF (Non-negative matrix factorization) method was pioneered by Lee and Seung \cite{lee1999learning} to decompose a non-negative matrix into two non-negative matrices. Moreover, since the dimensions of the factorizing matrices are much smaller than those of the original matrix, the NMF can also be considered as a low rank approximation method just like the PCA and the SVD. However, the strength of NMF lies in its ability to better infer the latent structure of the data. This is what makes it a powerful technique for solving complex problems. Moreover, NMF has a relationship with k-means clustering \cite{huang2021robust}, spectral clustering \cite{lu2014non}, probabilistic latent semantic indexing (the most used technique in information retrieval) \cite{ding2006nonnegative}  and sparse data transformation \cite{hoyer2004non}. Thus, it can be effectively adapted to many problems in machine learning, in general and data clustering, data representation and dimensionality reduction, in particular. Usually, performing this adaptation consists of adding a new term(s) to the original NMF cost function to impose certain constraints on the factorizing matrices and/or adding some penalty terms to direct the optimization process toward intended objective. For instance, \cite{cai2010graph} adds to NMF a neighborhood graph based penalty term to preserve the pairwise closeness of the data points in the reduced space; similarly, \cite{ahmed2021neighborhood} adds a term to preserve the geometric data structure in the reduced space which is modeled by a minimum spanning tree calculated from the neighborhood similarity graph; \cite{ding2006orthogonal} adds terms to enforce orthogonality of the factorizing matrices to perform data clustering in the reduced space;  \cite{HEDJAM2021107814} adds a term to preserve a feature relationship between the original data and the reduced data in order to freely adjust the centroids of computed clusters. Although adding a penalty term to NMF has been shown to be very beneficial, controlling balancing factor (a hyper-parameter) between the NMF cost function and the added term still remains a problem. Usually, the hyper-parameter is either adjusted experimentally or calculated analytically (e.g. using Lagrangian multipliers). Let's define these kind of NMF-based methods as parameterized NMF methods.

Moreover, the original NMF and the NMF-based methods mentioned above adopt unsupervised approaches to estimate the factorizing matrices. Thus, the ability to perform prediction (e.g., classification) using the new dimensions (i.e., features) is not guaranteed. To solve this problem, some works have been proposed to supervise the NMF by including the data labels when calculating the factorizing matrices. There are two categories of supervised NMF methods. The first category uses the principle of linear discriminant analysis to improve the prediction performance in reduced space  \cite{xue2006modified, MA2021107676}, while the second category integrates the data labels in the loss function  \cite{leuschner2019supervised, peng2021robust}. 

In this article, in-line with the second category methods, we propose a new supervised learning NMF-based framework. The motivation is that through the literature, NMF has been shown to be an outstanding method for extracting relevant, useful and meaningful reduced non-negative features and, if supervised by including labels in the factorization process, can be used as a good predictor, or can be used to successfully assist subsequent classifiers make better predictions.
Therefore, we want to demonstrate that it is possible to learn the best factorizing matrices (i.e., coefficient matrix or the latent bases that represent the training data, and the basis matrix in which the data are represented) by finding the best hyper-parameter for a parameterized NMF method using a training-validation framework (supervised learning) empowered by the Genetic algorithm and the majority voting rule. As a result, the learned basis matrix will be used as a projection matrix to transform future samples (test samples) for subsequent use. In the training phase, the goal is to estimate the optimal hyper-parameter along with the factorizing matrices. Based on the new representation of training data, classifiers can be built and learned. In the testing phase, the test data is first represented by the latent bases via solving a pseudo-inverse problem. Then it is fed to the learned classifier to predict its label. To validate the concept described above, two parameterized NMF methods are analyzed, GNMF \cite{cai2010graph}, and FR-NMF \cite{HEDJAM2021107814}.

In addition to the framework described above, we have also proposed a new NMF-based combination learning algorithm, designed to solve a multi-class classification problem. The proposed algorithm is based on the hypothesis that supervised NMF works better with fewer classes. In other words, NMF can produce better latent bases when the number of classes is small. The experiments conducted in this work support this hypothesis; i.e., the higher the number of classes, the lower the classification performance and vice versa. Therefore, we proposed to define the optimal factorizing matrices along with the hyper-parameter by applying a parameterized NMF algorithm separately to different class-pairs instead of applying it once to the whole dataset. The proposed algorithm consists of the following steps: First, divide the main dataset into multiple subsets, where each subset is related to two classes. Note here that the order of the classes in each class-pair does not matter. Second, estimate two factorizing matrices and one hyper-parameter for each class-pair. Third, since a unique parameter may not be suitable to all class-pairs, use Genetic Algorithm to identify the best combination of hyper-parameters that provides the best overall classification performance. A GA can approximates the solution of numerically complex problems faster and more efficiently. It has been used to solve complex problems such as clustering and forecasting scenarios \cite{maulik2000genetic}. This method searches for viable solutions in a particular area. Simply, a new generation with updated objective values is extracted from a randomly generated candidate. This algorithm is repeated until a satisfactory result is obtained. The prediction of the labels of the training data and the validation of the classification performance are carried out according to a cross-validation and majority voting scheme which will be described in more detail in Section \ref{seq:method}. To our knowledge, no similar learning framework has been proposed in the literature.  

\section{NMF and some parameterized NMF-based methods}
\label{seq-nmf}

Formally, let $\textbf{X}\in \mathbb{R}^{m\times n}$ be a matrix of $n$ columns representing the nonegative samples and $m$ rows representing their features, and $r<\{m,n\}$. NMF aims to find non-negative matrices $\textbf{W}\in\mathbb{R}^{m\times r}$ and $\textbf{H}\in\mathbb{R}^{r\times n}$ that minimize the following cost-function:

\begin{equation}
f(\textbf{W}, \textbf{H}) = \frac{1}{2}\|\textbf{X}-\textbf{W}\textbf{H}\|_F^2,
\label{eq-nmf}
\end{equation}
where $\|_.\|_F^2 $ represents the Frobenius norm. The model in Eq. (\ref{eq-nmf}) can also be formulated as an optimization problem of the form:
\textcolor{black}{
	\begin{equation}
	\min_{\text{W},\text{H}>0}  \| \textbf{X}-\textbf{W}\textbf{H} \|_F^2 = \min_{\text{W},\text{H}>0}\sum_{i,j}(\textbf{X}-\textbf{W}\textbf{H} )_{ij}^2,
	\label{eq-optim1}
	\end{equation}
}
Using the multiplicative update rules for non-negative optimization system proposed by Lee and Seung \cite{lee1999learning}, $\textbf{H}$ and $\textbf{W}$ are \textcolor{black}{updated} by:

\begin{equation}
\textbf{H}^{(t+1)}\leftarrow \textbf{H}^{(t)} \odot \frac{\textbf{W}^{(t)^\top}\textbf{X}}{\textbf{W}^{(t)^\top}\textbf{W}^{(t)}\textbf{H}^{(t)}},
\label{eq-h}
\end{equation}

\begin{equation}
\textbf{W}^{(t+1)}\leftarrow \textbf{W}^{(t)} \odot \frac{\textbf{X}\textbf{H}^{(t+1)^\top}}{\textbf{W}^{(t)}\textbf{H}^{(t+1)}\textbf{H}^{(t+1)^\top}},
\label{eq-i}
\end{equation}
\textcolor{black}{where} $\odot$ stands for the element-wise matrix product, and $\frac{A}{B}$ stands for the element-wise matrix division. $\textbf{H}^{(0)}$ and $\textbf{W}^{(0)}$ are set to random values 
and the updates are repeated until \textbf{W} and \textbf{H} become stable.

\subsection{Graph regularized NMF (GNMF)}
The GNMF method \cite{cai2010graph} adds, to the cost function of NMF, a penalty term based on a similarity graph to preserve the neighborhood structure of the data points in the reduced space. The contribution of the penalty term is weighted by a hyper-parameter $\lambda$. The resulting model is therefore suitable for clustering on a manifold. GNMF minimizes the fllowing  objective function:

\begin{equation}
\min_{\text{W},\text{H}>0}\|\textbf{X}-\textbf{WH}\|^2_F + \lambda \text{Tr}(\textbf{HLH}^\top),
\label{eq-gnmf}
\end{equation}
where $\textbf{L}$ is called the Laplacian matrix computed as $\textbf{L}=\textbf{D}-\textbf{W}$, and $\textbf{D}$ is a diagonal matrix with $\textbf{D}_{jj}=\sum_l\textbf{W}_{jl}$. The derivation of this optimization problem leads to the following updating rules:

\begin{equation}
\textbf{W}^{(t+1)}\leftarrow \textbf{W}^{(t)} \odot \frac{\textbf{X}\textbf{H}^{(t)^\top}}{\textbf{W}^{(t)}\textbf{H}^{(t)}\textbf{H}^{(t)^\top}},
\label{eq-ii}
\end{equation}

\begin{equation}
\textbf{H}^{(t+1)}\leftarrow \textbf{H}^{(t)} \odot \frac{\textbf{X}^\top\textbf{W}^{(t+1)}+\lambda \textbf{W}^{(t+1)}\textbf{H}^{(t)^\top}}{\textbf{H}^{(t)^\top}\textbf{W}^{(t+1)^\top}\textbf{W}^{(t+1)}+\lambda \textbf{D} \textbf{H}^{(t)^\top}}.
\label{eq-h3}
\end{equation}

\subsection{Feature relationship-preservation NMF (FR-NMF)}

FR-NMF \cite{HEDJAM2021107814} adds, to the NMF cost function, a penalty term that is equivalent to imposing an orthogonality constraint on the coefficient matrix $\textbf{H}$. The two terms of the model are then balanced by a hyper-parameter $\lambda$ to allow a scale relationship between the scatter of data points and that of cluster centroids. FR-NMF minimizes the following objective function:

	\begin{align}
	\min_{W,H\geq 0}\frac{1}{2}\|\textbf{X}-\textbf{WH}\|_F^2 + \frac{1}{2}\|\textbf{XX}^\top - \lambda\textbf{WW}^\top\|_F^2, 
	\label{eq-model0}
	\end{align} where $\textbf{W}\in\mathbb{R}^{m\times k}$ and $\textbf{H}\in\mathbb{R}^{k\times n}$ are respectively the basis and coefficient matrices, and $\lambda$ is a positive scalar to be defined by experiment in this work. The multiplicative update rule leads to:

	\begin{equation}
	\textbf{H}^{(t+1)}\leftarrow \textbf{H}^{(t)} \odot \frac{{\textbf{W}^{(t)}}^\top\textbf{X}}{{\textbf{W}^{(t)}}^\top\textbf{W}^{(t)}\textbf{H}^{(t)}},
	\label{eq-ph}
	\end{equation}
	\begin{equation}
	\textbf{W}^{(t+1)} \leftarrow \textbf{W}^{(t)} \odot \frac{\textbf{X}{\textbf{H}^{(t+1)}}^\top+\lambda \textbf{XX}^\top \textbf{W}^{(t)}}{ \textbf{W}^{(t)}\textbf{H}^{(t+1)} {\textbf{H}^{(t+1)}}^\top+\frac{\lambda^2}{2}\textbf{W}^{(t)}\textbf{W}^{(t)^\top} \textbf{W}^{(t)} }.
	\label{eq-pw}
	\end{equation}

\section{Proposed learning framework}
\label{seq:method}

First, let's define some formal notations that will be used throughout the paper:

\begin{itemize}
	\item[-] $\textbf{X} \in \mathbb{R}^{d\times n}$: a matrix of $n$ samples of $d$ features each.
	\item[-] $m$: number of classes. 
	\item[-] $C_{i=1..m}$: the $i^{th}$ class.
	\item[-] $\textbf{y}\in \{0, 1,..,m\}^n$: label vector of the samples in $\textbf{X}$.
	\item[-]$\textbf{x} \in \mathbb{R}^{d\times 1}$: a given sample.
	\item[-] $y$: class label of $\textbf{x}$.
	\item[-]$\{S_{t=1..T}\}$: $T$ possible subsets, where each subset is composed of samples from two classes $C_i$ and $C_t$; $i\neq j$; i.e., $K=m(m-1)/2$.
	\item[-]$p_t$: the parameter of the NMF-based algorithm when applied to $S_t$.
	\item[-]$(p_1,..,p_t,..p_T)$: a chromosome of parameters.
	\item[-] $Pop$: a population of chromosomes.
\end{itemize}

As mentioned in the introduction section, in this work, two parameterized NMF algorithms are considered, FR-NMF \cite{HEDJAM2021107814} and GNMF \cite{cai2010graph}. The goal is to fine-tune, for each algorithm, its hyper-parameter so that we can generate the best factorizing matrices with relevant latent data for subsequent classification tasks. More specifically, learning the optimal parameter is achieved through a training-validation-test process. During the training/validation phase, the task is to estimate the optimal hyper-parameter along with the factorizing matrices; i.e., coefficient matrix (latent bases that represent the training data), and the basis matrix (new space in which the data are represented). Based on the new representation of training data, classifiers can be built and learned. The $k$-$NN$ (k-Nearest Neighbor with k=1) algorithm is used in this letter. Other classifiers like SVM, MLP, pre-trained CNN, will be investigated in a future more extended work. The optimal hyper-parameter obviously leads to maximizing the classification accuracy of the validation set. The cross-validation algorithm is used to estimate the true classification accuracy. In the testing phase, the test data is first represented by the latent bases via solving a pseudo-inverse problem, and is then fed to the learnt classifier to predict its label.  

Two factors can affect the effectiveness of any matrix factorization method, including NMF: the number of classes available and the overlap between classes. In other words, the calculation of the basis vectors of the reduced space is less complex with fewer less-overlapped classes, the thing we can validate by the results obtained from the different experiments carried out in this work. This motivated us to propose a new NMF learning framework in which, instead of training a single NMF-based algorithm for all classes at once, it would be better to train one on each class-pair $ S_t $ and decide the label of the validation (and test) samples based on a combination process (e.g. majority voting rule). Considering that the parameterized NMF algorithm uses one hyper-parameter, applying it to each $S_t$ involves learning a vector of $K$ hyper-parameters, $K$ basis matrices $\textbf{W}$ , and $K$ coefficient matrices $\textbf{H}$. In other words, for each subset $S_t$, the process requires learning $p_t$, $\textbf{W}_t$, and $\textbf{H}_t$. Since the distributions of the subset are not necessarily similar, $ p_{t=1..T} $ may not necessarily have the same value. Consequently, the matrix factorization of each subset leads to different factorizing matrices and therefore to different values of $ p_t $. Identifying the optimal hyper-parameter vector $ P = (p_1, .., p_t, .., p_T) $ is a complex problem and cannot be solved analytically (e.g., combinatorial if $ p_t $ are integers). We therefore propose to solve it by using a meta-heuristic optimization algorithms. In our work, the Genetic Algorithm is used. \\

To summarize, the learning framework consists of the following steps:
\subsection{Learning of the hyper-parameter vector and factorizing matrices}
\label{seq:learning-hyper}

\begin{itemize}
	\item[1)] divide the training part of $\textbf{X}$ into training/validation folds.
	\item[2)] each training fold is divided into K subsets $S = \{S_{t=1..T}\}$, where each $S_t$ is composed of two classes, $C_i$ and $C_t$, with $i\neq j$. The total number of subsets is equal to $\frac{m!}{2(m-2)!}$. The division by 2 is due to the fact that the order of the classes in each subset does not matter. 
	\item[3)] for each subset $S_t$, compute $\textbf{W}_t$ and $\textbf{H}_t$ using the NMF-based algorithm considered fed by its parameter $ p_t $.
	\begin{equation}
		\textbf{W}_t, \textbf{H}_t\leftarrow AlgoNMF(S_t, p_{k}),
	\end{equation}
	where $AlgoNMF()$ is FR-NMF or GNMF. The vector $P = (p_1, .., p_t, .., p_T)$ is initialized randomly.
	\item[4)] for each sample $\textbf{x}$ in the validation fold, compute the corresponding latent sample (data) by solving pseudo-inverse problem:
	\begin{equation}
		\textbf{{h}}\leftarrow (\textbf{W}_t^\top \textbf{W}_t)^{-1}\textbf{W}^\top_t\textbf{x},
		\label{eq-hh}
	\end{equation}
	then predict its label using the k-Nearest Neighbor ($k$-$NN$) algorithm based on $\textbf{H}_t$:
	\begin{equation}
	 l \leftarrow kNN(\textbf{h}, \textbf{H}_t, \textbf{y}_t ).
	 \label{eq-knn}
	\end{equation}
	 In this letter we have used $1$-$NN$ classifier. Therefore, the label of the first nearest sample from $\textbf{H}_t$ to $\textbf{h}$ is assigned to $\textbf{x}$.
	\item[5)] Since 3) and 4) are repeated for all the $S_t, \forall t=1..T$, the process ends up with a list $L_t$ of labels for the same sample $\textbf{x}$. The final label of $\textbf{x}$ can be identified via a majority voting rule; i.e., the dominant label will be assigned to it.
	\item[6)] Compute the accuracy $acc$ for the current validation fold.
	\item[7)] Repeat 2) to 6) for all the training/validation folds, and compute the average accuracy $acc_{avg}$.	
\end{itemize}

The process from 1) to 7) is performed for one specific parameter vector $P=(p_1,..,p_t,..,p_T)$. The optimal parameter vector $\hat{P}$ that maximizes the average accuracy $acc_{avg}$ is searched using the Genetic algorithm. Since $P$ is a real-valued vector, the continuous GA is used. The parameter setting of GA will be described in Section \ref{seq:param}. We assume that the best factorizing matrices $ \hat{\textbf{H}}_t $ and $ \hat{\textbf{W}}_t; \forall t=1..T $ are those corresponding to $\hat{P}$. Algorithm 1 describes the training phase of the proposed framework. 

\subsection{Testing phase}
\label{seq:testing}

The prediction of the label of a given test sample is carried out in the same way as for the validation sample during the training phase. The optimal $ \hat{\textbf{H}}_t $ and $ \hat{\textbf{W}}_t ; \forall t=1..T$ will be used to predict the list of labels for this test sample by applying Eqs. (\ref{eq-hh}) and (\ref{eq-knn}), and the majority voting rule is then applied to decide on its  final label as describe in step 5) above.

\RestyleAlgo{boxruled}
\LinesNumbered
\begin{algorithm}[!htp]\small
\caption{Supervised Class-pairwise Parameterized NMF}
	\KwInput{\\
$~~~ - (\textbf{X}, \textbf{y})$: labeled dataset, with features scaled in [0,1]\;
		}
	\KwOutput{\\
$~~~ - \hat{P} = (\hat{p}_{1},..,\hat{p}_{k},..,\hat{p}_{K}])$: optimal parameter chromosome\;
$~~~ - \{( \hat{ \textbf{W} }, \hat{ \textbf{H} }  )_{t=1..T}\}$: set of optimal factorizing matrices of subsets\;
		}
	\textbf{initialization}\;
	$Acc_{chromo}\leftarrow []$ $\#$\textit{empty list of chromosome accuracies (fitnesses)}\;
	\textit{Initialize the population of chromosomes $Pop$ randomly}\;
	\While{$acc_{avg}$ doesn't improve \textbf{or} max itration is not achieved}{
	\For{each chromosome $P=(p_1,..,p_t,..p_T)$ in $Pop$}{
		$Acc_{valid} \leftarrow []$ $\#$ \textit{empty list of accuracies of validation folds}\;
		$\#$ \textit{Cross-validation data split}\;
		\For {each cross-validation split}{
			$\#$ \textit{\textbf{training}: estimate factorizing matrices using current training fold}\;
			\For {each $S_t$ }{
			$\textbf{W}_t, \textbf{H}_t\leftarrow AlgoNMF(S_t, p_{k})$\;
		}
			
			$\#$\textit{ \textbf{validation}: using validation fold}\;
			$T$: \textit{list of true labels of current validation fold}\;
			$L\leftarrow []$: \textit{empty list of predicted labels of current validation fold}\;
			\For {each sample \textbf{x} in curreent validation fold}{
				$L_t\leftarrow []$ $\#$ \textit{list of labels of $\textbf{x}$}\;
				\For{each $(\textbf{H}, \textbf{W})_t$}{			
				$\#$ \textit{compute the latent data \textbf{{h}} that represents \textbf{{x}}}\; 
				$\textbf{\textit{h}}\leftarrow (\textbf{W}_t^\top \textbf{W}_t)^{-1}\textbf{W}^\top_t\textbf{\textit{x}}$ $\#$ \textit{pseudo-inverse reconstruction}\;
				$\#$ \textit{predict the label of \textbf{h}}\;
				$l\leftarrow kNN(\textbf{\textit{h}}, \textbf{H}_t, \textbf{y}_t)$\;
				$L_t.append(l)$\;			
			}
				$\#$\textbf{ majority voting labeling}\;
				$l \leftarrow MajVoting(L_t)$\;
				$L.append(l)$				
		}

		$acc \leftarrow Accuracy(L, T)$\;
		$Acc_{valid}.append(acc)$
		}
		$\#$ \textit{compute average accury for all validation folds}\;
		$acc_{avg} \leftarrow average(Acc_{valid}) $\;
		$Acc_{chromo}.append(acc_{avg})$
	}
Chromosome selection step\;
Chromosome cross-over step\;
Chromosome mutation step
}
The chromosomes of the final population (generation) are sorted according to their fitnesses, and the chromosome with the highest fitness is selected as the optimal, ie., $\hat{P} =(\hat{p}_{1},..,\hat{p}_{k},..,\hat{p}_{K})$. The corresponding $\{( \hat{ \textbf{W} }, \hat{ \textbf{H} }  )_{t=1..T}\}$ are selected as well.
\end{algorithm}

\begin{figure*}[!ht]\scriptsize
	\centering
	\begin{tabular}{c}
		\includegraphics[scale=0.5]{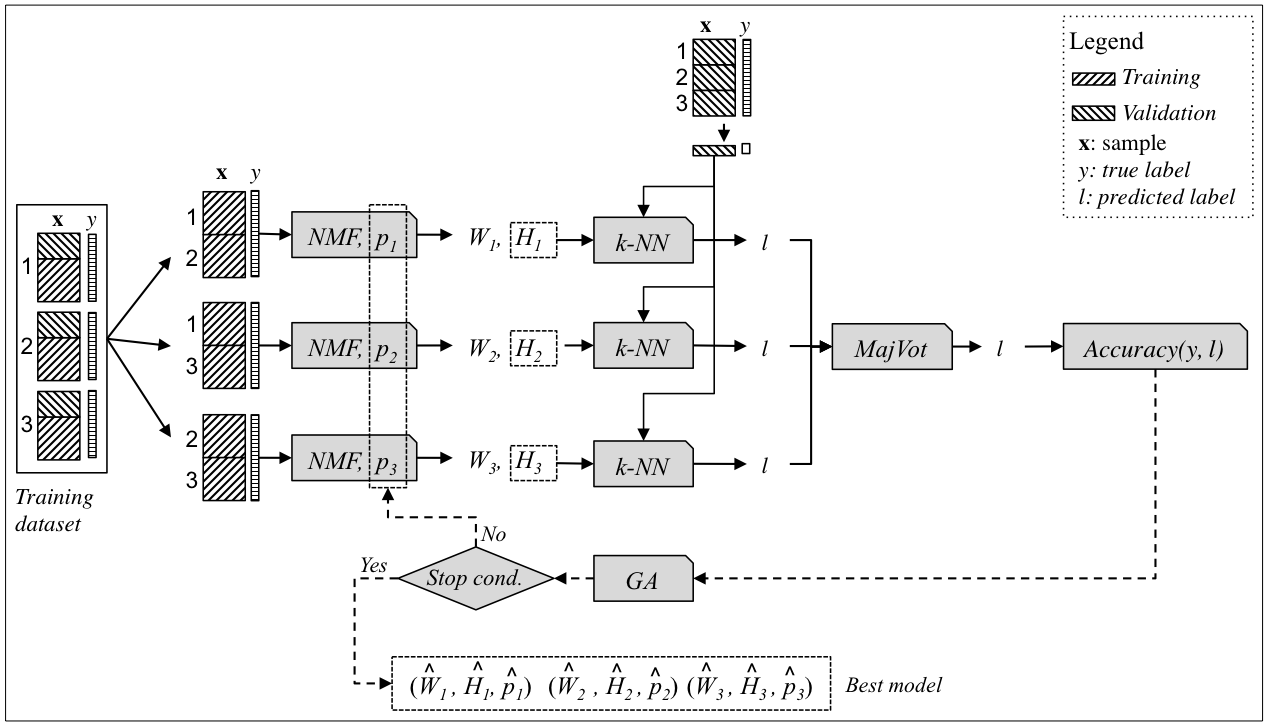}
	\end{tabular}
	\caption{Class-pairwise parameterized NMF learning framework (cNMF). Example of 3-class problem.}
	\label{fig-cnmf}
\end{figure*}

Figure \ref{fig-cnmf} illustrates the overall training phase of the proposed parameterized NMF learning framework. 

\section{Experimental results and evaluation}
\label{seq:experiments}

The proposed supervised class-pairwise NMF learning framework (named cNMF) is evaluated on real-life and synthetic datasets using two parameterized NMF algorithms (FR-NMF and GNMF). To demonstrate its effectiveness, two experiments are carried out. In the first experiment, the parameterized NMF algorithms are trained and tested based on all classes at once. Let's name this strategy as unique parameterized NMF (uNMF). In the second experiments they are trained and tested on each class-pair separately as described above. First let's describe the datasets considered, and the parameter setting of each framework part.  

\subsection{Data description}
\label{sq-datasets}
Four (4) real-life datasets are used to evaluated the performance of the framework. The first dataset is the Columbia Object Image Library (COIL20)\footnote[1]{\url{https://www.cs.columbia.edu/CAVE/software/softlib/coil-20.php}}. It consists of 1440 images of 20 objects (with 72 poses each). Each image is of 128$\times$128 grayscale pixels leading to 1056784 features. The second dataset is the Olivetti face dataset (Oface), which contains 400 face images divided into 40 classes. Each class holds 10 face images of one distinct subject. Each image is of $64\times 64$ pixels leading to 4096 features. To reduce the time complexity we selected 60 best features for COIL20 and Oface datasets using SlectKBest algorithm\footnote[3]{\scriptsize\url{https://scikit-learn.org/stable/modules/classes.html##module-sklearn.feature_selection}}. The third dataset is the Optical Recognition of Handwritten Digits dataset, which consists of 1797 images divided into 10 classes. The size of the images is $8\times 8$ pixels leading to 64 features each. The fourth dataset is the Wine dataset, which consists of 178 samples divided into 3 classes. The dimension of the samples is 13 real-valued features. The 2nd, 3rd and 4th datasets can be downloaded here\footnote[3]{\scriptsize\url{https://scikit-learn.org/stable/modules/classes.html##module-sklearn.datasets}}.

The last set of synthetic datasets consists of 15 datasets generated using the \texttt{make\_blob} module from SciKit Python library. Blob-like datasets are controlled by the number of clusters ($cl$), the number of features ($f$) and the standard deviation of clusters ($\sigma$). Each dataset contains 1000 samples and $\sigma$ is set to 1. For reproducibility purposes, the Python statements to generate Blobs datasets can be found in the footnote below\footnote[4]{\scriptsize\texttt{datasets.make\_blobs(center\_box=(1,5),n\_samples=1000,centers=$c$,n\_features=$f$, cluster\_std=1,random\_state=0)}}. Table \ref{t-data} describes the datasets considered in this work in terms of number of classes ($cl$) and number of features ($f$).

\begin{table}[!htb]\scriptsize
	\caption{Description of the datasets considered. }
	\label{t-data}
	\centering
	\begin{tabular}{|l|c|c|c|c|c}
		\cline{1-5}
		\textbf{Datasets} & \textbf{Wine}  & \textbf{Digits} & \textbf{COIL20} & \textbf{Ofaces} & \multicolumn{1}{l}{}                 \\ \cline{1-5}
		$cl$                 & 3              & 10              & 20              & 40              & \multicolumn{1}{l}{}                 \\ \cline{1-5}
		$f$                 & 13             & 64              & 60              & 60              & \multicolumn{1}{l}{}                 \\ \hline \hline
		\textbf{Datasets} & \textbf{C3F10} & \textbf{C4F10}  & \textbf{C6F10}  & \textbf{C8F10}  & \multicolumn{1}{c|}{\textbf{C10F10}} \\ \hline
		$cl$                 & 3              & 4               & 6               & 8               & \multicolumn{1}{c|}{10}              \\ \hline
		$f$                 & 10             & 10              & 10              & 10              & \multicolumn{1}{c|}{10}              \\ \hline \hline
		\textbf{Datasets} & \textbf{C3F20} & \textbf{C4F20}  & \textbf{C6F20}  & \textbf{C8F20}  & \multicolumn{1}{c|}{\textbf{C10F20}} \\ \hline
		$cl$                 & 3              & 4               & 6               & 8               & \multicolumn{1}{c|}{10}              \\ \hline
		$f$                 & 20             & 20              & 20              & 20              & \multicolumn{1}{c|}{20}              \\ \hline \hline
		\textbf{Datasets} & \textbf{C3F40} & \textbf{C4F40}  & \textbf{C6F40}  & \textbf{C8F40}  & \multicolumn{1}{c|}{\textbf{C10F40}} \\ \hline
		$cl$                 & 3              & 4               & 6               & 8               & \multicolumn{1}{c|}{10}              \\ \hline
		$f$                 & 40             & 40              & 40              & 40              & \multicolumn{1}{c|}{40}              \\ \hline
	\end{tabular}
\end{table}

\subsection{Parameter setting}
\label{seq:param}

The parameters of the framework are: the parameter vector $(p_1, p_2,..,p_T)$ and the GA operators (selection, cross-over and mutation). The range of the parameters $p_t; \forall t$ is set to $[0..1]$ (continuous chromosomes). The chromosomes selection method used is the selection by tournament, the cross-over method is the one point mate method with a probability set to 0.2, the mutation probability is set to 0.05, the population size is set to 10 chromosomes, and the number of generations (iterations) is set to 20. In this work, the best GA setting is defined experimentally using the DEAP (Distributed Evolutionary Algorithms in Python) library. The description of the GA operators used in this work can be found here\footnote[5]{\texttt{https://deap.readthedocs.io/en/master/api/tools.html\#operators}}.  

\subsection{Experimental results}
\label{seq:results}
 For each dataset and for each parameterized NMF algorithm (FR-NMF or GNMF), we have i) applied cNMF and uNMF for the following lower ranks $ r = [2, 4, 6, 8, 10] $, ii) applied the classifier kNN to each lower ranked dataset, and finally iii) calculated the mean classification accuracy ($meanAcc$) over all the ranks considered. Tables \ref{tab-gnmf-rd} and \ref{tab-frnmf-rd} show $meanAcc$ of cNMF and uNMF for GNMF and FR-NMF respectively using the real-life datasets, and Tables \ref{tab-gnmf-sd} and \ref{tab-frnmf-sd} show the same using the synthetic datasets. We can see that cNMF outperforms uNMF in all the cases. In some datasets like Digits, the mean accuracy gap is remarkable. This means that applying FR-NMF or GNMF according to the cNMF strategy identifies relevant latent bases that best represent data in a reduced space and, therefore,  leading to better classification performance.

\begin{table}[!htb]\tiny
		\caption{Classification average accuracy ($meanAcc$) for GNMF on real-life datasets.}
	\label{tab-gnmf-rd}
	\centering
	\begin{tabular}{l|c|c|c|c|c|c|c|c|}
		\cline{2-9}
		& \multicolumn{2}{c|}{\textbf{COIL20}} & \multicolumn{2}{c|}{\textbf{Digits}} & \multicolumn{2}{c|}{\textbf{OFaces}} & \multicolumn{2}{c|}{\textbf{Wine}} \\ \hline
		\multicolumn{1}{|l|}{Method} & cNMF         & uNMF         & cNMF         & uNMF         & cNMF         & uNMF         & cNMF        & uNMF        \\ \hline
		\multicolumn{1}{|l|}{$meanAcc$} & 0.79         & 0.74         & 0.92         & 0.75         & 0.61         & 0.50         & 0.94        & 0.94        \\ \hline
	\end{tabular}
\end{table}

\begin{table}[!htb]\scriptsize
		\caption{Classification average accuracy ($meanAcc$) for GNMF on synthetic datasets.}
	\label{tab-gnmf-sd}
	\centering
	\begin{tabular}{|c|c|c|c|c|c|c|}
		\hline
		$f$       & \multicolumn{2}{c|}{\textbf{10}} & \multicolumn{2}{c|}{\textbf{20}} & \multicolumn{2}{c|}{\textbf{40}} \\ \hline
		\textbf{methods} & \textbf{cNMF}   & \textbf{uNMF}  & \textbf{cNMF}   & \textbf{uNMF}  & \textbf{cNMF}   & \textbf{uNMF}  \\ \hline
		$cl= 3$            & 0.77            & 0.71           & 0.99            & 0.95           & 1.00            & 0.99           \\ \hline
		$cl= 4$            & 0.70            & 0.59           & 0.99            & 0.96           & 1.00            & 0.95           \\ \hline
		$cl= 6$            & 0.63            & 0.55           & 0.98            & 0.92           & 1.00            & 0.93           \\ \hline
		$cl= 8$            & 0.67            & 0.57           & 0.97            & 0.84           & 1.00            & 0.91           \\ \hline
		$cl= 10$           & 0.70            & 0.51           & 0.95            & 0.80           & 1.00            & 0.87           \\ \hline
	\end{tabular}
\end{table}

\begin{table}[!htb]\tiny
		\caption{Classification accuracy ($meanAcc$) for FR-NMF on real-life datasets.}
	\label{tab-frnmf-rd}
	\centering
	\begin{tabular}{l|c|c|c|c|c|c|c|c|}
		\cline{2-9}
		& \multicolumn{2}{c|}{\textbf{COIL20}} & \multicolumn{2}{c|}{\textbf{Digits}} & \multicolumn{2}{c|}{\textbf{OFaces}} & \multicolumn{2}{c|}{\textbf{Wine}} \\ \hline
		\multicolumn{1}{|l|}{\textbf{Method}} & cNMF         & uNMF         & cNMF         & uNMF         & cNMF         & uNMF         & cNMF        & uNMF        \\ \hline
		\multicolumn{1}{|l|}{$meanAcc$} & 0.58         & 0.43         & 0.89         & 0.63         & 0.52         & 0.41         & 0.92        & 0.93        \\ \hline
	\end{tabular}
\end{table}

\begin{table}[!htb]\scriptsize
		\caption{Classification accuracy ($meanAcc$) for FR-NMF on synthetic datasets.}
	\label{tab-frnmf-sd}
	\centering
	\begin{tabular}{|l|c|c|c|c|c|c|}
		\hline
		\multicolumn{1}{|c|}{$f$}       & \multicolumn{2}{c|}{\textbf{10}}                                        & \multicolumn{2}{c|}{\textbf{20}}                                        & \multicolumn{2}{c|}{\textbf{40}}                                        \\ \hline
		\multicolumn{1}{|c|}{\textbf{methods}} & \multicolumn{1}{c|}{\textbf{cNMF}} & \multicolumn{1}{c|}{\textbf{uNMF}} & \multicolumn{1}{c|}{\textbf{cNMF}} & \multicolumn{1}{c|}{\textbf{uNMF}} & \multicolumn{1}{c|}{\textbf{cNMF}} & \multicolumn{1}{c|}{\textbf{uNMF}} \\ \hline
		$cl= 3$                                  & 0.77                               & 0.80                               & 0.99                               & 0.95                               & 1.00                               & 1.00                               \\ \hline
		$cl= 4   $                               & 0.76                              & 0.71                              & 0.99                              & 0.96                              & 1.00                                  & 0.95                              \\ \hline
		$cl= 6$                                  & 0.70                              & 0.67                              & 0.98                              & 0.92                             & 1.00                                  & 0.97                              \\ \hline
		$cl= 8$                                  & 0.73                               & 0.67                               & 0.97                               & 0.88                              & 1.00                              & 0.90                              \\ \hline
		$cl= 10$                                 & 0.69                              & 0.58                              & 0.92                              & 0.80                              & 1.00                              & 0.84                              \\ \hline
	\end{tabular}
\end{table}

\begin{figure}[!ht]\scriptsize
	\centering
	\begin{tabular}{cc}
		\includegraphics[scale=0.22]{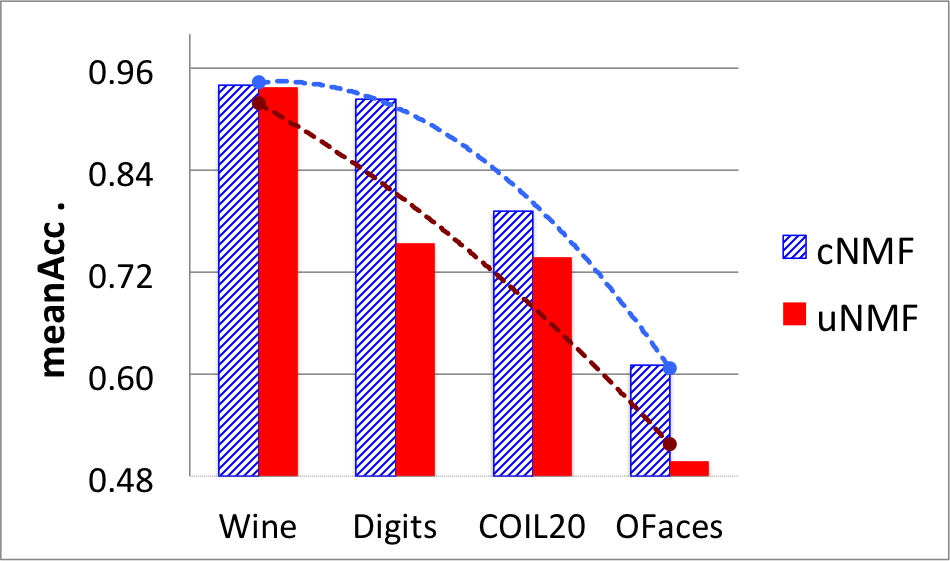}&\hspace{-0.25cm}
		\includegraphics[scale=0.22]{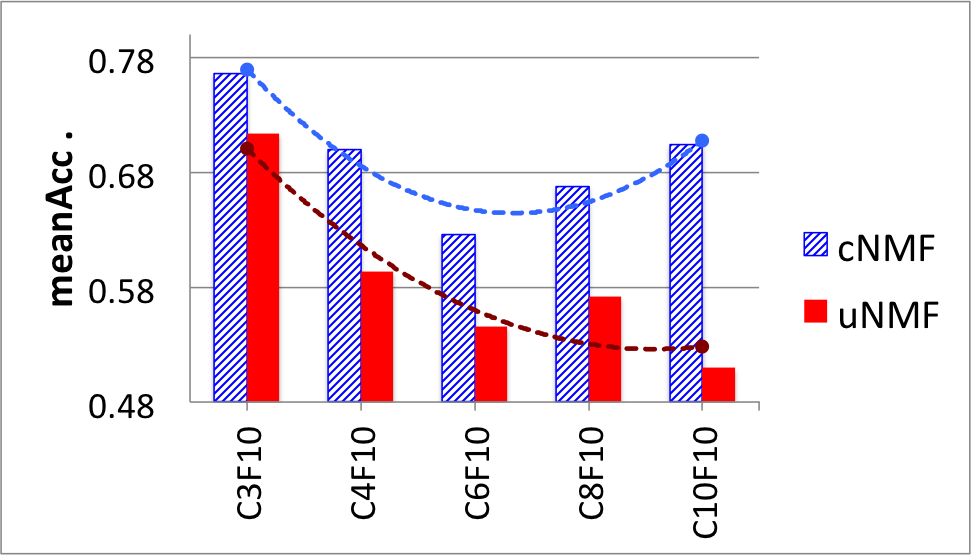}\\
		(a) & (b)\\
		\includegraphics[scale=0.22]{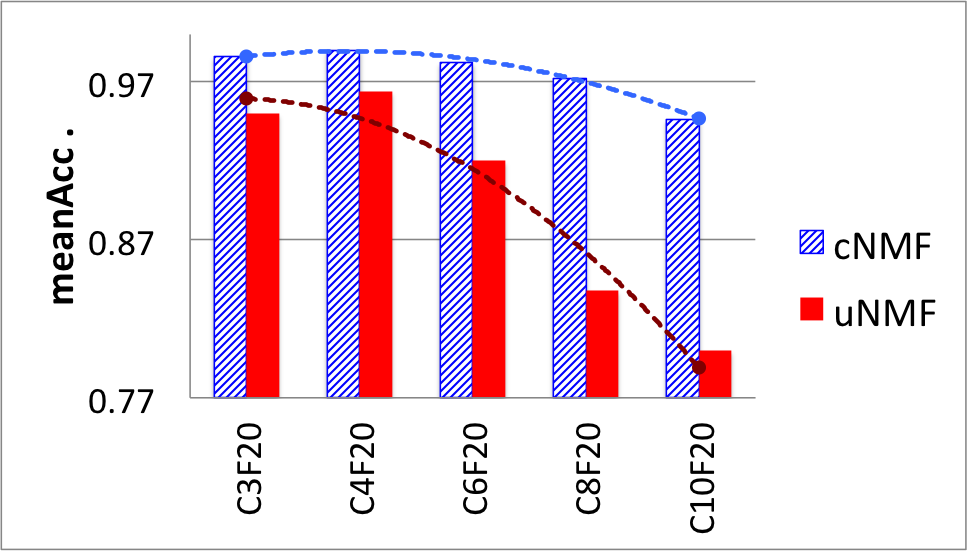}&\hspace{-0.25cm}
		\includegraphics[scale=0.22]{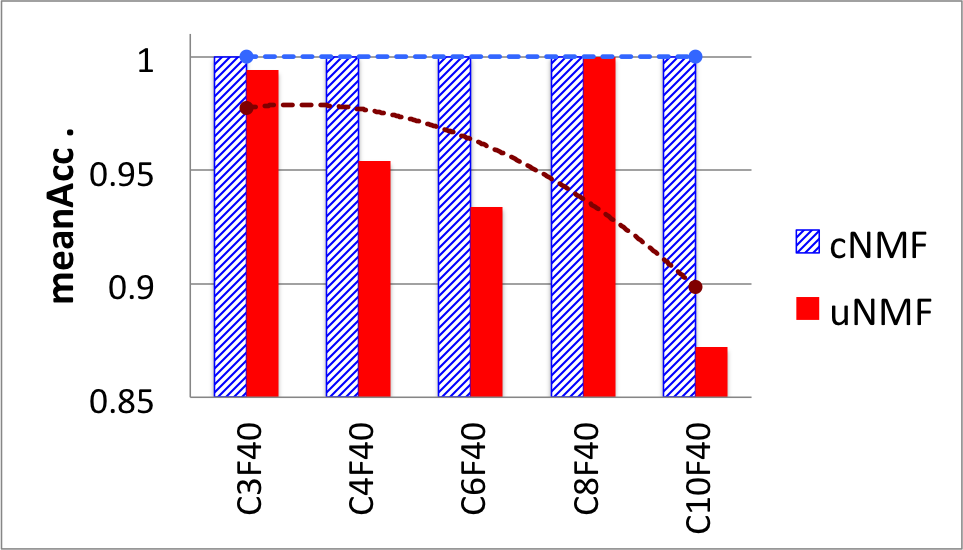}\\
		(c) & (d)
	\end{tabular}
	\caption{Graphical representation of the $meanAcc$ of GNMF on real-life datasets (a) and synthetic datasets (b-d). Trend lines (dotted) are added for better analysis.  }
	\label{fig-gnmf}
\end{figure}

\begin{figure}[!ht]\scriptsize
	\centering
	\begin{tabular}{cc}
		\includegraphics[scale=0.22]{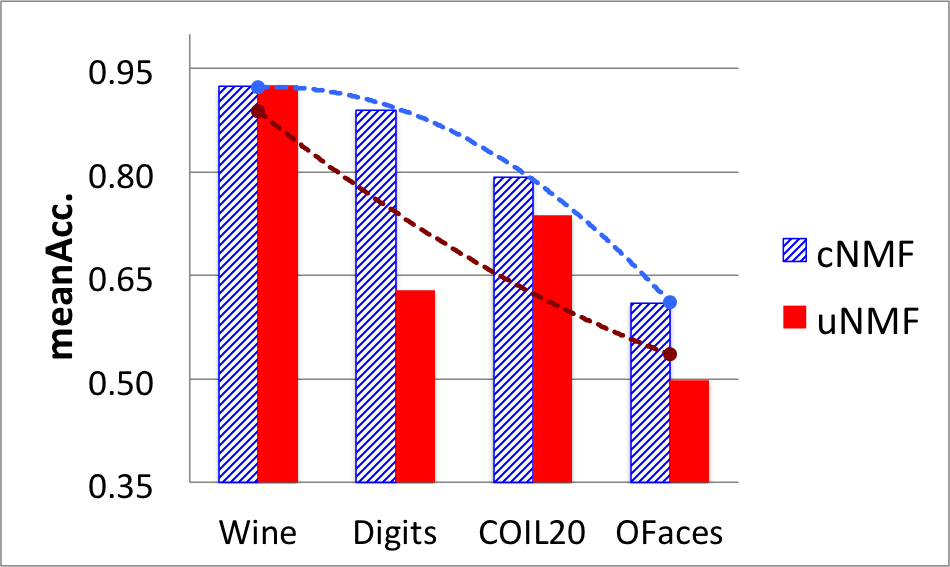}&\hspace{-0.25cm}
		\includegraphics[scale=0.22]{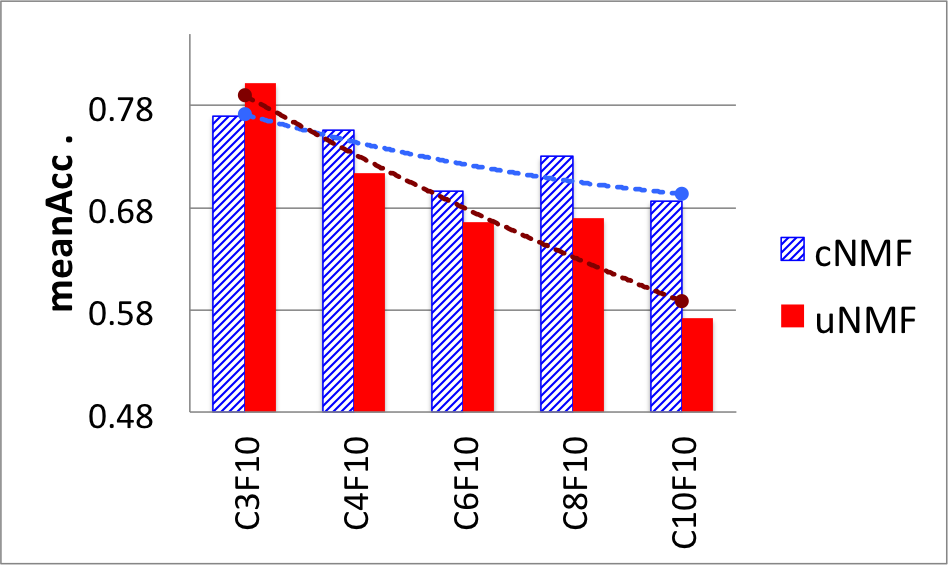}\\
		(a) & (b)\\
		\includegraphics[scale=0.22]{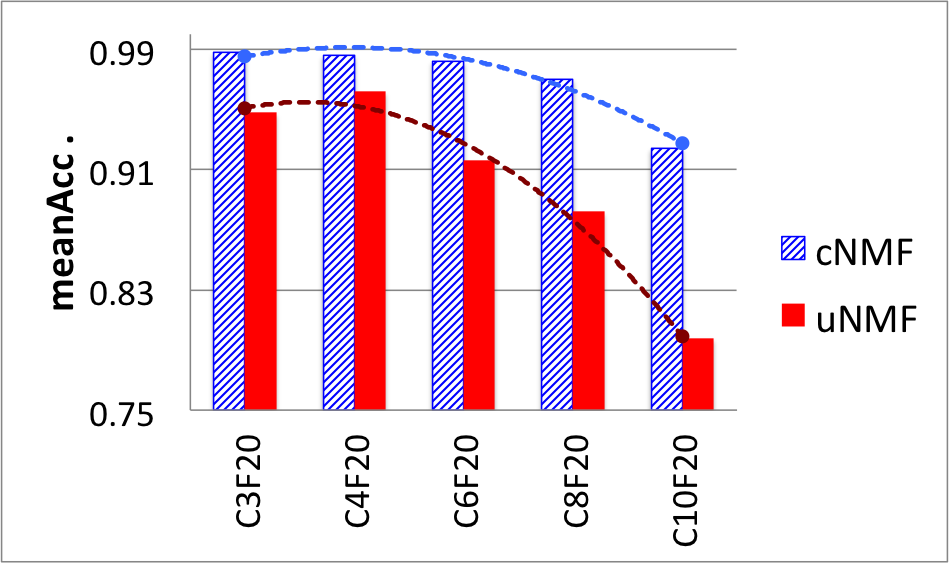}&\hspace{-0.25cm}
		\includegraphics[scale=0.22]{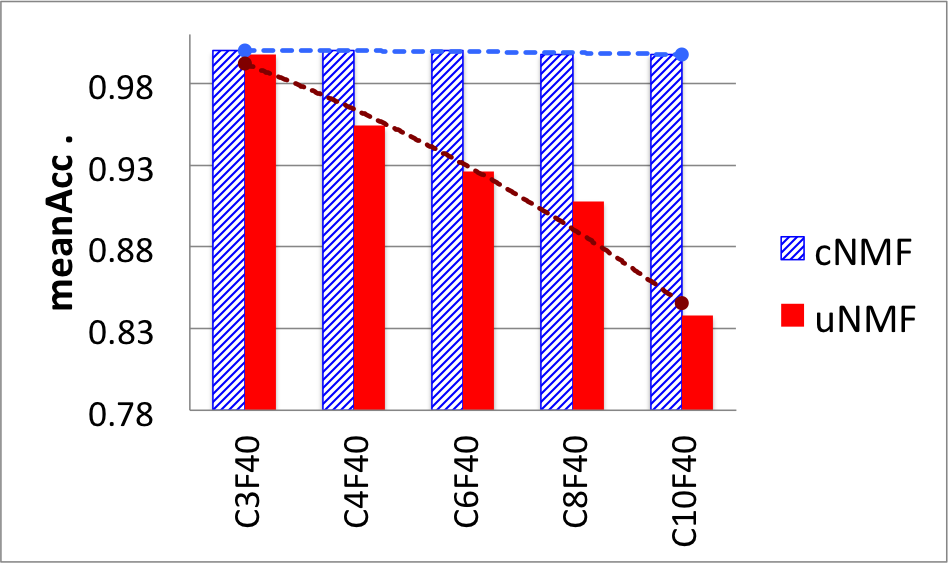}\\
		(c) & (d)
	\end{tabular}
	\caption{Graphical representation of the $meanAcc$ of FR-NMF on real-life datasets (a) and synthetic datasets (b-d). Trend lines (dotted) are added for better analysis. }
	\label{fig-frnmf}
\end{figure}

For visual analysis we plotted the $meanAcc$ of cNMF and uNMF for each dataset as shown in Figures \ref{fig-gnmf} and \ref{fig-frnmf}. In order to visually simplify the analysis of the results in each graph, datasets are sorted according to the number of classes they contain from left (smallest) to right (largest). From the figures, the following observations can be drawn:
\begin{itemize}
	\item[i)] in general, the classification performance according to uNMF strategy is inversely proportional to the number of classes in a dataset. This is well indicated by the rapid decrease of the red bars in each graph (from left to right). This finding supports our hypothesis stated in Section \ref{seq:method} regarding the the impact of the number of classes on the matrix factorization task.
	\item[ii)] cNMF is less affected than uNMF by the number of classes. The corresponding $meanAcc$ bars show a smooth decrease from left to right in each graphic. In some cases the performance is not influenced at all (see graphics (d) in Figures \ref{fig-gnmf} and \ref{fig-frnmf}).  This means that applying parameterized NMF algorithms according to cNMF (class-pairwise) strategy improves the performance of the subsequent classification tasks.
	\item[iii)] for synthetic datasets the classification performance improves as the number of data features increases. This maybe due to the well defined data distribution (i.e., Gaussian).
	\item[iv)] on average, the performance of cNMF and that of uNMF on Wine dataset seem to be similar. However, cNMF performs better on small rank values $r$ while uNMF outperforms cNMF for large values of $r$. Another possible reason is because Wine dataset has only three classes with few number of samples.
\end{itemize}

\begin{figure}[!ht]\scriptsize
	\centering
	\begin{tabular}{cc}
		\includegraphics[scale=0.18]{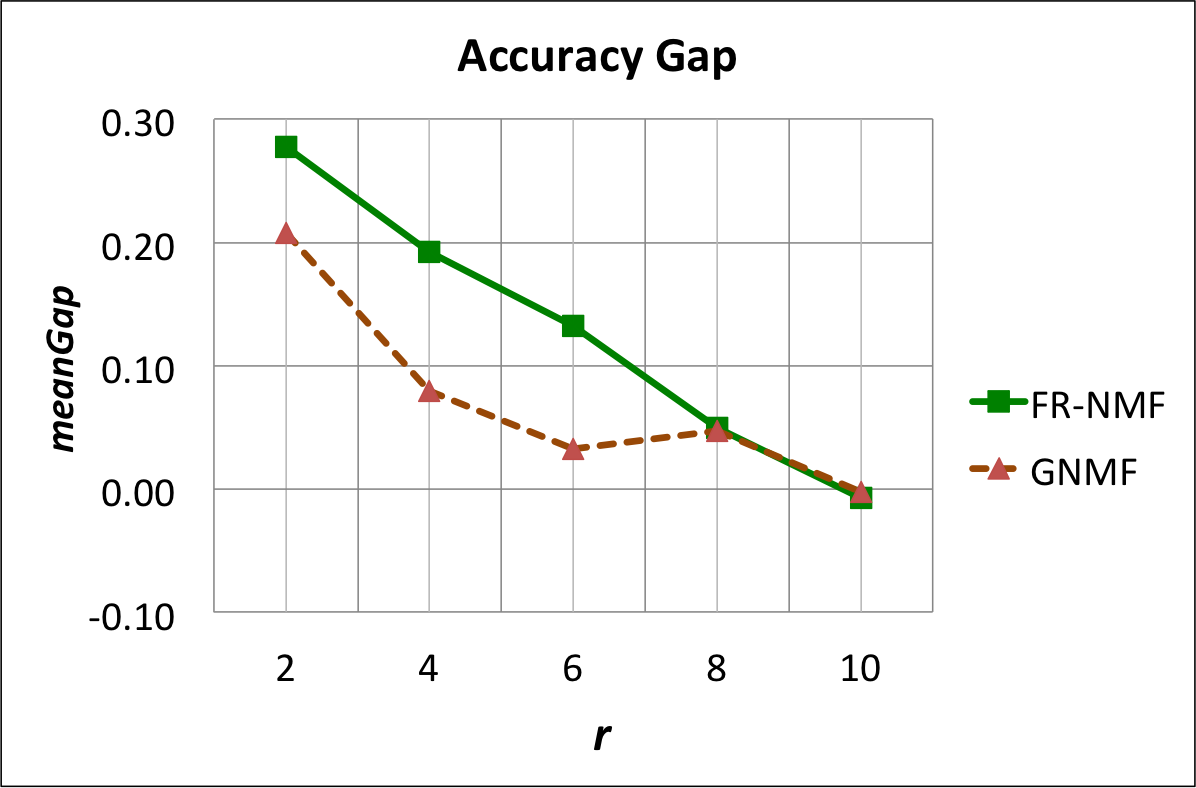}&\hspace{-0.25cm}
		\includegraphics[scale=0.18]{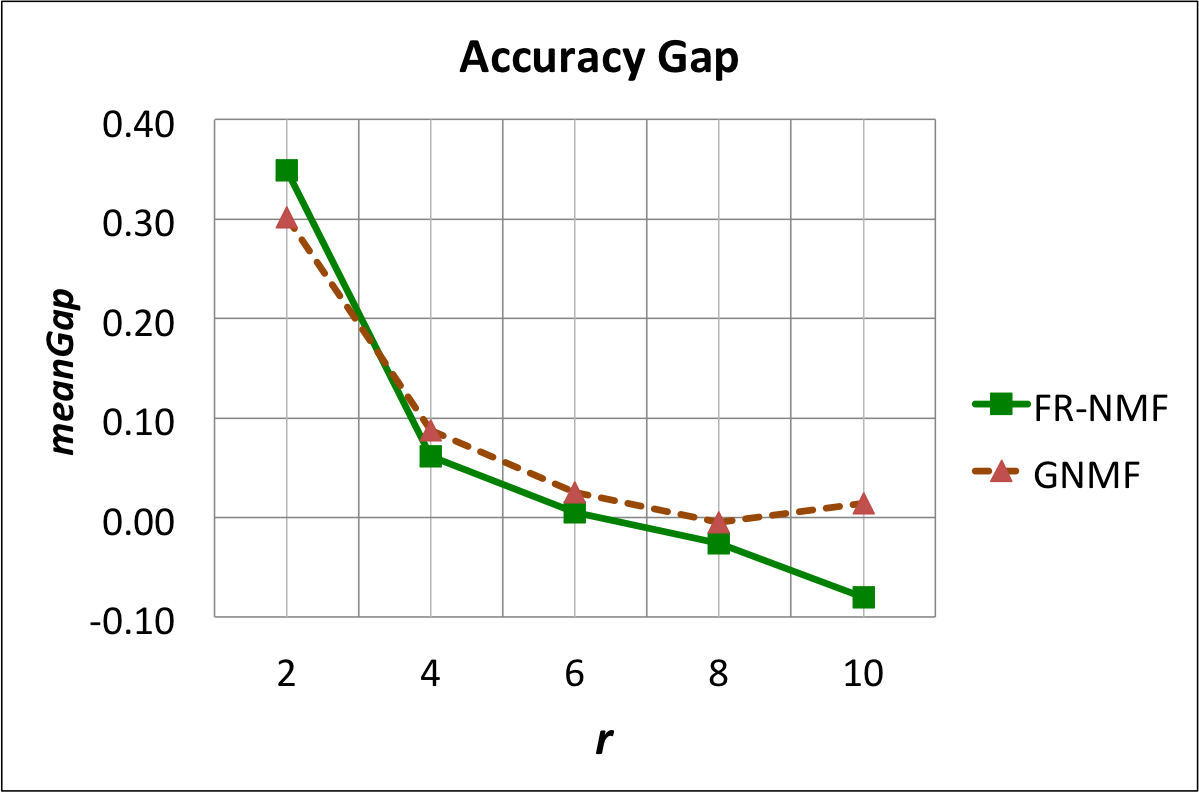}\\
		(a) & (b)
	\end{tabular}
	\caption{Classification accuracy gap ($meanGap$) between cNMF and uNMF strategies on real-life datasets (a) and on synthetic datasets (b). The dotted line represent the trending lines. }
	\label{fig-gap}
\end{figure}

In order to understand in more details the difference between the two strategies, cNMF and uNMF, we investigated the classification accuracy gap between them for each low rank value $r$. Let's define $Gap = meanAcc_{cNMF} - meanAcc_{uNMF}$ for one dataset, and let's define $meanGap$ as the average gap over all the datasets considered. Figures \ref{fig-gap} (a) and (b) show $ meanGap $ on real-life and synthetic datasets respectively for each row $ r $. We can notice that cNMF greatly outperforms uNMF for the smallest $ r $. A significant gap of $35\%$ for $r=2$ is obtained important. This indicates the usefulness of the proposed approach in particular for applications requiring smaller data dimension (e.g., data visualization). 

\begin{figure}[!ht]\scriptsize
	\centering
	\begin{tabular}{cc}
		\includegraphics[scale=0.18]{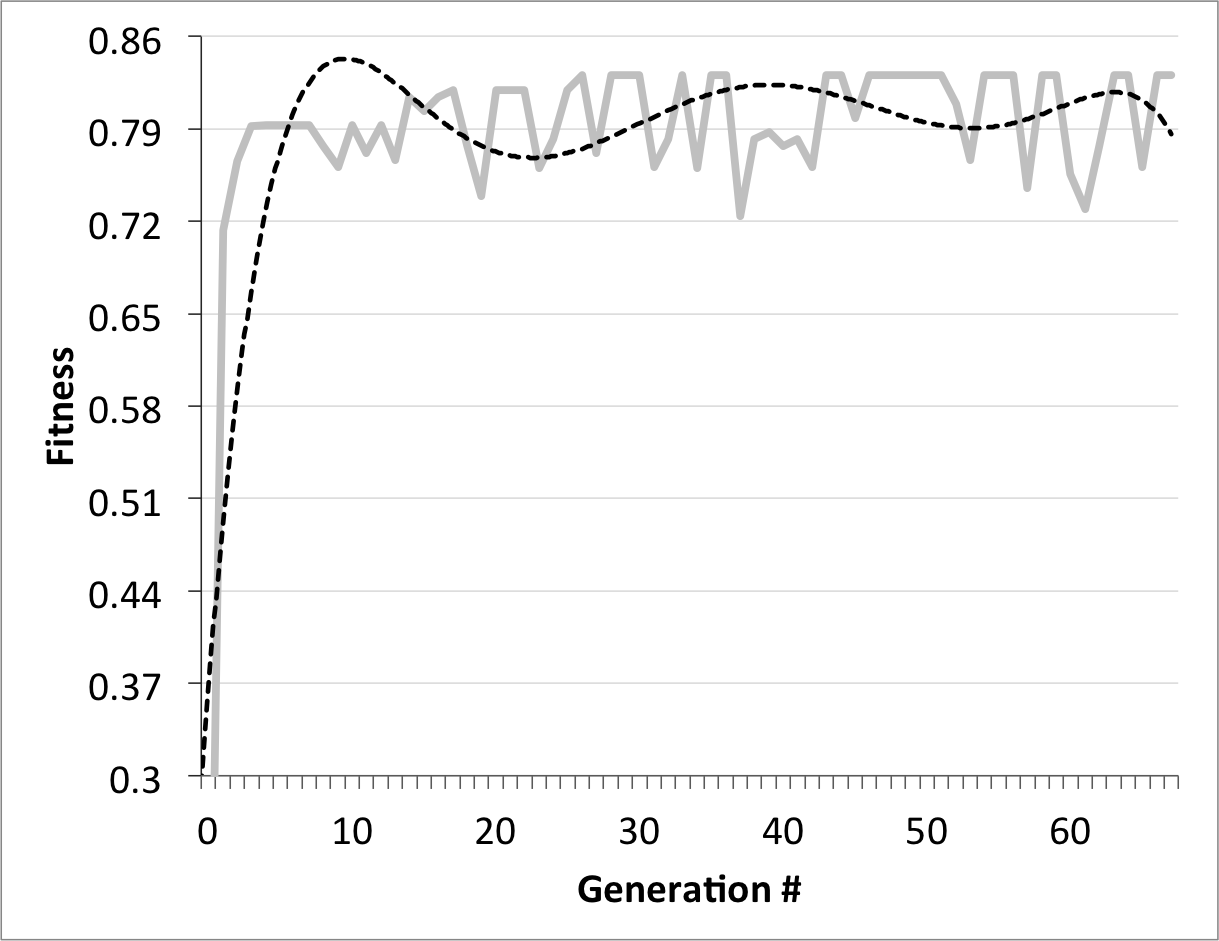}&\hspace{-0.25cm}
		\includegraphics[scale=0.18]{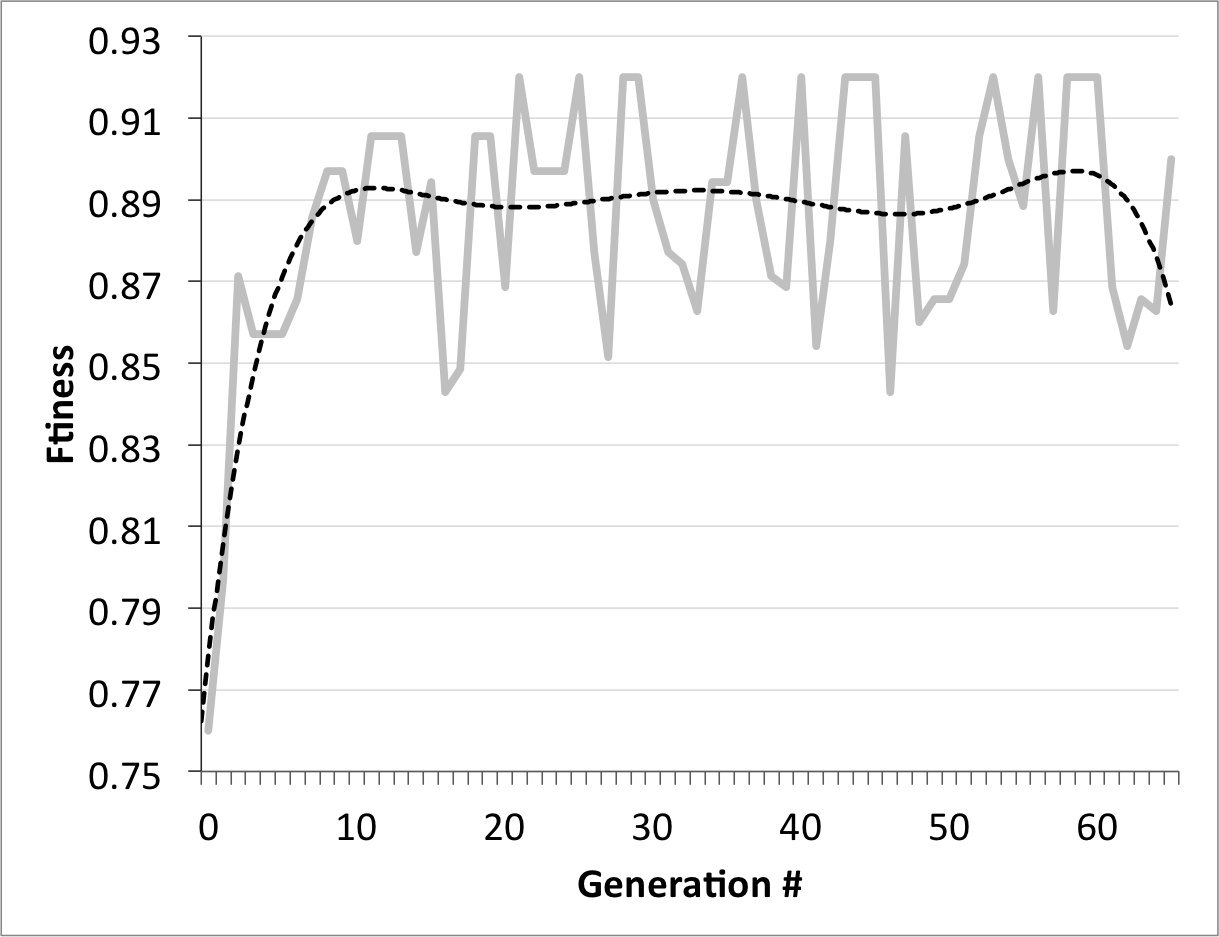}\\
		(a) & (b)
	\end{tabular}
	\caption{GA evolution. Fitness vs number of generations on (a) real-life dataset (e.g., Digits) and on (b) synthetic dataset (e.g., C6F20).}
	\label{fig-ga}
\end{figure}

Figure \ref{fig-ga} shows the evolution of fitness with the number of generations of the genetic algorithm on (a) a real-life dataset (e.g., Digits) and (b) a synthetic dataset (e.g., C6F20). We can see that over generations, new relevant populations are sought after and the fitness improves. Experimentally, we have found that fitness does not change significantly after 20 generations for all the datasets considered in this work.

\section{Conclusion}
\label{seq-conclusion}

This work presents a novel learning framework designed to adapt any parameterized NMF algorithm (NMF with additional term controlled by a parameter) for classification problems. Given a labeled dataset, the basic idea is to train a parameterized NMF algorithm in the aim of identifying the optimal value of the parameter. This  value should produce factorizing matrices that maximize the subsequent classification tasks. However the main novelty of the proposed framework consists of training the parameterized NMF algorithm  in an evolutionary manner on each class-pair separately and then combining the predicted labels using the combining scheme described in Section \ref{seq:method} and illustrated in Figure \ref{fig-cnmf}.  This was based on the hypothesis that the matrix factorization process is impacted by the number of classes.  The higher the number the classes is the more difficult to identify the factorizing matrices. We have validated this hypothesis experimentally and found that the new combination learning strategy is more effective. In order to demonstrate the effectiveness of the class-pairwise NMF learning strategy, we considered two parameterized NMF algorithms, GNMF and FR-NMF. In other words, we have applied each of them according to the uNMF strategy where each algorithm is applied on the entire dataset (all classes) at once, and also according to cNMF where it is applied to each pair of classes, and finally we have analyzed the difference between the two strategies in terms of classification performance obtained based on real-life and synthetic datasets. The strategy cNMF is found more effective. The advantage of the framework is that it may use: i) any parameterized NMF algorithm and be extended to learning more than one parameter, ii) any kind of classifier (instead of $k$NN) to optimize the parameter and the factorizing matrices, iii) any kind of label aggregation strategy (instead of majority voting rule). We believe that NMF, as an outstanding feature extractor, when supervised and trained, can discover latent data more suitable for classification tasks. It is also easy to implement and understand (as illustrated in Figure \ref{fig-cnmf}). The downside is that a higher number of classes makes the framework learning process more time demanding because it requires repeating the learning process $ \frac {m!} {2 (m-2)!} = \frac{m(m-1)}{2} $ times. 

In a more extended future work, we will integrate supervised parameterized NMF with other classifiers like SVM, MLP and CNN, and evaluate them on other kinds of datasets. We will also evaluate the framework using other parameterized NMF methods.

%
%

 \section*{Acknowledgements}
\footnotesize{The author would like to thank the SQU Internal Grant (IG/SCI/COMP/22/04) for their financial support.}

\bibliographystyle{unsrtnat}
\bibliography{references}  

\begin{thebibliography}{14}
\providecommand{\natexlab}[1]{#1}
\providecommand{\url}[1]{\texttt{#1}}
\expandafter\ifx\csname urlstyle\endcsname\relax
  \providecommand{\doi}[1]{doi: #1}\else
  \providecommand{\doi}{doi: \begingroup \urlstyle{rm}\Url}\fi

\bibitem[Lee and Seung(1999)]{lee1999learning}
Daniel~D Lee and H~Sebastian Seung.
\newblock Learning the parts of objects by non-negative matrix factorization.
\newblock \emph{Nature}, 401\penalty0 (6755):\penalty0 788--791, 1999.

\bibitem[Huang et~al.(2021)Huang, Kang, Xu, and Liu]{huang2021robust}
Shudong Huang, Zhao Kang, Zenglin Xu, and Quanhui Liu.
\newblock Robust deep k-means: An effective and simple method for data
  clustering.
\newblock \emph{Pattern Recognition}, 117:\penalty0 107996, 2021.

\bibitem[Lu et~al.(2014)Lu, Fu, and Shu]{lu2014non}
Hongtao Lu, Zhenyong Fu, and Xin Shu.
\newblock Non-negative and sparse spectral clustering.
\newblock \emph{Pattern Recognition}, 47\penalty0 (1):\penalty0 418--426, 2014.

\bibitem[Ding et~al.(2006{\natexlab{a}})Ding, Li, and
  Peng]{ding2006nonnegative}
Chris Ding, Tao Li, and Wei Peng.
\newblock Nonnegative matrix factorization and probabilistic latent semantic
  indexing: Equivalence chi-square statistic, and a hybrid method.
\newblock In \emph{AAAI}, volume~42, pages 137--43, 2006{\natexlab{a}}.

\bibitem[Hoyer(2004)]{hoyer2004non}
Patrik~O Hoyer.
\newblock Non-negative matrix factorization with sparseness constraints.
\newblock \emph{Journal of machine learning research}, 5\penalty0 (9), 2004.

\bibitem[Cai et~al.(2010)Cai, He, Han, and Huang]{cai2010graph}
Deng Cai, Xiaofei He, Jiawei Han, and Thomas~S Huang.
\newblock Graph regularized nonnegative matrix factorization for data
  representation.
\newblock \emph{IEEE transactions on pattern analysis and machine
  intelligence}, 33\penalty0 (8):\penalty0 1548--1560, 2010.

\bibitem[Ahmed et~al.(2021)Ahmed, Hu, Acharya, and Ding]{ahmed2021neighborhood}
Imtiaz Ahmed, Xia~Ben Hu, Mithun~P Acharya, and Yu~Ding.
\newblock Neighborhood structure assisted non-negative matrix factorization and
  its application in unsupervised point-wise anomaly detection.
\newblock \emph{Journal of Machine Learning Research}, 22\penalty0
  (34):\penalty0 1--32, 2021.

\bibitem[Ding et~al.(2006{\natexlab{b}})Ding, Li, Peng, and
  Park]{ding2006orthogonal}
Chris Ding, Tao Li, Wei Peng, and Haesun Park.
\newblock Orthogonal nonnegative matrix t-factorizations for clustering.
\newblock In \emph{Proceedings of the 12th ACM SIGKDD international conference
  on Knowledge discovery and data mining}, pages 126--135, 2006{\natexlab{b}}.

\bibitem[Hedjam et~al.(2021)Hedjam, Abdesselam, and Melgani]{HEDJAM2021107814}
Rachid Hedjam, Abdelhamid Abdesselam, and Farid Melgani.
\newblock Nmf with feature relationship preservation penalty term for
  clustering problems.
\newblock \emph{Pattern Recognition}, 112:\penalty0 107814, 2021.
\newblock ISSN 0031-3203.

\bibitem[Xue et~al.(2006)Xue, Tong, Chen, Zhang, and He]{xue2006modified}
Yun Xue, Chong~Sze Tong, Wen-Sheng Chen, Weipeng Zhang, and Zhenyu He.
\newblock A modified non-negative matrix factorization algorithm for face
  recognition.
\newblock In \emph{18th International Conference on Pattern Recognition
  (ICPR'06)}, volume~3, pages 495--498. IEEE, 2006.

\bibitem[Ma et~al.(2021)Ma, Zhang, and Zhang]{MA2021107676}
Jiaqi Ma, Yipeng Zhang, and Lefei Zhang.
\newblock Discriminative subspace matrix factorization for multiview data
  clustering.
\newblock \emph{Pattern Recognition}, 111:\penalty0 107676, 2021.
\newblock ISSN 0031-3203.

\bibitem[Leuschner et~al.(2019)Leuschner, Schmidt, Fernsel, Lachmund, Boskamp,
  and Maass]{leuschner2019supervised}
Johannes Leuschner, Maximilian Schmidt, Pascal Fernsel, Delf Lachmund, Tobias
  Boskamp, and Peter Maass.
\newblock Supervised non-negative matrix factorization methods for maldi
  imaging applications.
\newblock \emph{Bioinformatics}, 35\penalty0 (11):\penalty0 1940--1947, 2019.

\bibitem[Peng et~al.(2021)Peng, Ser, Chen, and Lin]{peng2021robust}
Siyuan Peng, Wee Ser, Badong Chen, and Zhiping Lin.
\newblock Robust semi-supervised nonnegative matrix factorization for image
  clustering.
\newblock \emph{Pattern Recognition}, 111:\penalty0 107683, 2021.

\bibitem[Maulik and Bandyopadhyay(2000)]{maulik2000genetic}
Ujjwal Maulik and Sanghamitra Bandyopadhyay.
\newblock Genetic algorithm-based clustering technique.
\newblock \emph{Pattern recognition}, 33\penalty0 (9):\penalty0 1455--1465,
  2000.

\end{thebibliography}






\end{document}